%% file: 0-main.tex
\pdfoutput=1

\documentclass[11pt]{article}

\usepackage{acl}

\usepackage{times}
\usepackage{latexsym}

\usepackage[T1]{fontenc}

\usepackage[utf8]{inputenc}

\usepackage{microtype}
\usepackage{nert}
\usepackage[noorphans,vskip=0.8ex]{quoting}
\usepackage[flushleft]{threeparttable}

\title{Sentence-level Feedback Generation for English Language Learners: \\ Does Data Augmentation Help?}

\author{
Shabnam Behzad\hspace*{40pt} \\
Georgetown University\hspace*{40pt}\\
\eml{shabnam@cs.georgetown.edu}\hspace*{40pt}
\And Amir Zeldes \qquad Nathan Schneider\\
Georgetown University\\
\{\emldisplay{amir.zeldes@georgetown.edu}{amir.zeldes}, \emldisplay{nathan.schneider@georgetown.edu}{nathan.schneider}\}\texttt{@georgetown.edu}}

\begin{document}
\maketitle
\begin{abstract}

In this paper, we present strong baselines for the task of Feedback Comment Generation for Writing Learning. Given a sentence and an error span, the task is to generate a feedback comment explaining the error. Sentences and feedback comments are both in English. We experiment with LLMs and also create multiple pseudo datasets for the task, investigating how it affects the performance of our system. We present our results for the task along with extensive analysis of the generated comments with the aim of aiding future studies in feedback comment generation for English language learners.

\end{abstract}

\section{Introduction}
\input{1-intro--related}

\section{Experiments}\label{lab:data}

\input{2-experiments}

\section{Results and Analysis}\label{sec:classification}
\input{3-analysis}

\section{Discussion}\label{sec:retrieval}
\input{4-discussion}

\section{Acknowledgements}
We thank the shared task organizers and anonymous reviewers for their insightful comments.
This research was supported in part by NSF award IIS-2144881.

\bibliography{anthology,custom}
\bibliographystyle{acl_natbib}

\end{document}

%% file: 1-intro--related.tex
Grammatical error correction has been vastly studied recently in the NLP community \cite{wang2021comprehensive}, but it is not always sufficient to merely provide the learner with a correction; in many cases, explicit feedback can facilitate the learning
process. Language learners can revise improperly employed linguistic elements by reviewing feedback containing information on the error such as an explanation of why the usage is incorrect and suggestions on how to correct it. This will also help the user avoid making
similar errors in the future~\cite{pilan2020dataset}.

In this paper, we focus on preposition errors made by English language learners. Some studies have shown that the majority of syntactic errors made by English language learners are prepositional errors of substitution, omission, and addition~\cite{lorincz2012difficulties}. Prepositions are challenging for language learners to master since they are highly frequent; short, unstressed and perceptually weak; and can have several different senses which may not map onto their native languages~\cite{tyler2003semantics,morimoto2007comparison,johansson2015linguistic}.

The task of feedback generation hasn't been explored much until recently when~\citet{nagata-2019-toward} proposed the feedback comment generation task and a corpus~\cite{nagata-etal-2020-creating} and then organized the GenChal~2022: FCG (Feedback Comment Generation for Writing Learning) shared task~\cite{nagata-etal-2021-shared}. In this task, a system generates an explanation note, given a sentence and a span that indicates the error in the sentence.

Later, \citet{hanawa-etal-2021-exploring,hanawa2022analyzing} explored different baselines for this task, including a neural-retrieval-based method, a pointer-generator-based seq2seq model, and a retrieve-and-edit method. For preposition-related errors, they found the pointer-generator-based seq2seq model performs the best.

In this paper, we describe our submission to GenChal~2022: FCG~\cite{nagata-etal-2021-shared}. We use a simple encoder-decoder model to tackle the task and provide extensive analysis of the different aspects of the task. Our contributions in this paper are as follows:
\begin{itemize}
   \item We present a simple but strong baseline for the FCG task which is currently ranked third on the leaderboard (team \emph{GU}, BLEU score 0.472; top leaderboard score is 0.486).
    \item We look into data augmentation techniques and their usefulness for this task.
    \item We analyze samples that were marked as incorrect by human evaluators and categorize the errors made by our system.
    \item We further investigate the automatic evaluation metric used for the task and whether or not it is in line with human evaluations.
    
\end{itemize}

%% file: 2-experiments.tex
\subsection{Data}
We use data provided by~\citet{nagata-etal-2021-shared}. The sentences come from essays in ICNALE \citep[The International Corpus Network of Asian Learners of English;][]{ishikawa2013icnale}. ICNALE contains essays on two topics: “It is important for college students to have a part-time job” and “Smoking should be completely banned at all the restaurants in the country”.

\citet{nagata-etal-2021-shared} hired annotators to annotate a subset of the data for preposition errors. Annotators manually annotated all preposition errors with feedback comments in Japanese~\cite{nagata-2019-toward} and later translated these comments to English for the FCG shared task. The corpus consists of 4868, 170, and 215 sentences in the train, dev, and test sets respectively. The input for the task is a sentence and a span of the text which contains the error. The output is a string that explains why the span is erroneous. Example:

\begin{quoting}
\noindent \textit{Input:} And we can put posters to remind the smokers the risks they are taking .	37:48	

\noindent \textit{Output: }When the <verb> <<remind>> is used to express `` to cause someone to remember something'', ``someone'' is an <object> and a <preposition> needs to precede ``something''. Look up the use of the <verb> <<remind>> in a dictionary and add the appropriate <preposition> according to the context.
\end{quoting}

If a sentence contains more than one preposition error,
it can appear more than once in the training set, each time with a different span offset. We incorporate span offsets by adding special characters before and after the erroneous span before encoding the text. For example, the above input sentence becomes: \textit{And we can put posters to remind the *** smokers the *** risks they are taking .} We do not perform any further preprocessing since the text is already processed and tokenized. We used *** as special characters in our setting but the tokenizer behaved the same way when using other characters such as \{.

\subsection{Experimental Setting}

As a baseline for this task, we use T5~\cite{2020t5} as our model. 
T5 is an encoder-decoder model built on top of the transformer architecture~\cite{vaswani2017attention} which is pretrained using a combination of masked language modeling and multitask training such as summarization, machine translation, and sentiment classification.

In our experiments, we encode the essay sentences and fine-tune the model to decode feedback comments. 
We fine-tune \textit{T5-Large} (770M parameters) with the following hyper-parameters: batch size = 8, learning rate = 0.0001 and maximum training epoch = 50\footnote{\url{https://github.com/shabnam-b/GU-FCG-2022}}.

\subsection{Pseudo Data}
We experiment with a few other settings, trying to leverage pseudo data. To create the pseudo data, we select random sentences that are in the same corpus as the gold data  \citep[an ICNALE subset that includes correction of sentences;][]{ishikawa2018icnale} but are not included in the FCG shared task train/dev/test sets. Since the focus of the FCG shared task is on preposition errors, we use ERRANT~\cite{felice-etal-2016-automatic,bryant-etal-2017-automatic} to annotate error types in these sentences. Then we keep the samples that have preposition-related errors. This gave us 544 additional sentences. To obtain comments for these new sentences, we use our fine-tuned T5 model and generate comments for these samples. We experiment with the pseudo data in two ways:

\paragraph{Multi-stage fine-tuning} Fine-tune T5 on pseudo data, and then fine-tune that model on gold training data.
\paragraph{Combined fine-tuning} Combine pseudo and gold data, and fine-tune T5 on the combination.

Other than experimenting with pseudo data created from the same learner corpus, we create a large pseudo dataset from other learner corpora, W\&I+LOCNESS~\cite{bryant-etal-2019-bea,granger2014computer}. W\&I (Write \& Improve) is an online web platform in which users from around the world submit letters, stories,
articles, and essays, and the system provides automated feedback. Some of these submissions have been further corrected by annotators. LOCNESS consists of essays written by native
British and American undergraduates on different topics. 

Using ERRANT, we select sentences from W\&I+LOCNESS that have preposition errors. This resulted in 6,973 sentences.
For the grammatical error correction task, \citet{kiyono-etal-2019-empirical} suggests that when the amount of pseudo data and gold data is balanced, concatenating them for training works better (combined fine-tuning), but when the amount of data is unbalanced, a multi-step approach works better (multi-stage fine-tuning). Here, we investigate this by comparing conditions where the pseudo data is limited to 5,000 samples (balanced) versus conditions with all 6,973 pseudo samples (unbalanced).

\begin{table*}
\centering
\normalsize
\begin{tabular}{l|c|c||c}
\hline
\textbf{Model} & \textbf{Dev BLEU} &\textbf{Test BLEU}&\textbf{Human Evaluation F1 (Test)}\\
\hline
FCG Shared Task Baseline & 46.30 & 33.40& 31.16 \\
F/t T5 Large (No pseudo data) & \textbf{57.29} &47.11&58.60\\
Multi-stage f/t (ICNALE) & 55.01 & 46.76&--\\
Combined f/t (ICNALE)  & 55.55 &\textbf{47.25}&\textbf{61.90}\\
Multi-stage f/t (WIL, balanced) & 55.46 &45.95 &--\\
Combined f/t (WIL, balanced)  & 57.05 &46.91&61.40\\
Multi-stage f/t (WIL, unbalanced) & 55.05 & 44.97&--\\
Combined f/t (WIL, unbalanced) & \textbf{57.29} &45.36&--\\

\hline
\end{tabular}
\caption{Comparison of models on dev and test sets. \textit{WIL} refers W\&I+LOCNESS. The gold training data on which T5 is fine-tuned contains 4,868 samples. The multi-stage fine-tuning and combined fine-tuning conditions make use of data augmentation, supplementing the gold training data with pseudo data. The pseudo data consists of 5,000 samples in the balanced setting and 6,973 samples in the unbalanced setting. There are 170 and 215 samples in the dev and test sets, respectively. Best scores in each column are bolded.}
\label{tab:results}
\end{table*}

%% file: 3-analysis.tex
Results of our experiments are available in \cref{tab:results}. We compared against the official shared task baseline system, which was an encoder-decoder with a copy mechanism based on a pointer generator network. 

\subsection{Automatic Evaluation}

We use test set BLEU scores to compare all the conditions in \cref{tab:results}.
On this metric, all systems based on T5 give improvements of 12+ points over the official baseline.
The gain for our best model (which uses pseudo data from ICNALE for combined fine-tuning) is almost 14 points. 

\paragraph{Multi-stage vs.~combined fine-tuning}
In all our experiments, \textit{Combined f/t} showed better performance compared to \textit{Multi-stage f/t} (by a difference of 1 BLEU point or less).

\paragraph{Balanced vs.~unbalanced}

 In our experimental setup, using a larger pseudo dataset hurt the performance in both \textit{Combined f/t} and \textit{Multi-stage f/t} settings. One possible explanation is the amount of noise that is being introduced to the system by pseudo data. Creating pseudo data with different techniques might show different results.

\paragraph{In-domain vs.~out-of-domain pseudo data}
Even though our in-domain pseudo data was very small (544 sentences), it was more effective than larger amounts of out-of-domain pseudo data. An intuitive explanation for this case is that ICNALE contains essays on only two specific topics: “It is important for college students to have a part-time job” and “Smoking should be completely banned at all the restaurants in the country”. Since the FCG shared task test set comes from ICNALE, a more general model fine-tuned on pseudo data from other corpora might not necessarily perform well on this test set. It seems likely that the model trained on multiple datasets would be more robust in realistic settings testing on other domains.

\subsection{Human Evaluation}

\begin{figure}
\centering
\includegraphics[scale=.47, trim = 3cm 8cm 3cm 8cm, clip]{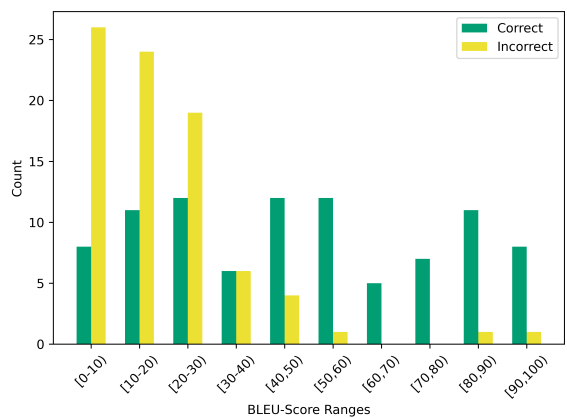}

\caption{Comparison of human evaluations (correct or incorrect system generated feedback comment) with automatic evaluation metric (BLEU score)}
\label{fig:eval}
\end{figure}

Shared task organizers provided us with the human evaluation of three of our systems (4th column in~\cref{tab:results}). In this evaluation, each system output is compared to the corresponding reference. System output is regarded as appropriate if the following criteria are met: A) it contains information similar to the reference and B) it does not contain information that is irrelevant to the erroneous span. The performance is measured by recall, precision, and F1 based on correct/incorrect outputs\footnote{\url{https://fcg.sharedtask.org/task/}}.

Based on this human evaluation, our best model achieved an F1 score of 61.09 (this was not our official submission to the shared task, but falls just behind the top leaderboard score\footnote{As of 14 December 2022} of 62.15). Comparing the performance of different systems, human evaluation results and test set BLEU scores seem to be consistent. We investigate this further for our top system, comparing human labels (correct or incorrect) with the BLEU score for each sample in the test set. Results are available in~\cref{fig:eval}. Based on this analysis, when BLEU score is higher than 60\%, it is mostly in line with the human evaluations. We also observe that about 49 samples (23\% of the test set) are indeed correct, but get a BLEU score below 50. This is due to system-generated comments not having much overlap with the gold feedback comment, despite being correct~\cite{sulem-etal-2018-bleu,nema-khapra-2018-towards}.

Lastly, we look at 50\% of samples where the model-generated comment was labeled as incorrect in human evaluation. We observed that generated comments are very fluent and follow the templates FCG annotators used. In cases where the system output was labeled as incorrect, some of the patterns we observed are as follows:

\textit{Completely incorrect comment ($\approx$54\%):} The model's generated comment includes incorrect suggestions and explanations (first and second example in~\cref{tab:example}). Interestingly, we noticed that the model made the same wrong suggestion in different sentences, containing the same type of error (for example, usage of “on” when it means \textit{sticking to, or hanging from a surface} such as “on the door”, “on the wall”). Possible explanations for these cases are that 1)~similar errors were not seen during training and 2)~in most cases, the sentence contains other errors within the same span or nearby tokens, which presumably makes it hard for the model to understand what the learner was trying to say.

\textit{Correct explanation, but incorrect suggestion ($\approx$22\%)}: In these cases, the model gives the right correction, but the explanation is incorrect or incomplete (third example in~\cref{tab:example}).

\textit{Correct suggestion, but incorrect evaluation ($\approx$14\%)}: In many cases, the model gives the correct suggestion but the comment starts with something along the lines of ``It is not grammatically incorrect to use the ...'', even though the usage is indeed incorrect (fourth example in~\cref{tab:example}).

\textit{Human annotation errors ($\approx$12\%):} In a few cases, we believe the system-generated comment is correct, but wrongly labeled as incorrect.

Looking at some positive examples, there are many cases where the model generates completely valid comments. In 19\% of cases (41 samples), the model generates exactly the same comment as the reference. In all of these instances, the exact comment was seen during training. There were another 51 comments in the test set that were seen during training, and the model was able to generate a correct comment (but not exactly the same) in 38 cases of those. In many cases, the system output has minor differences compared to the gold output but there are also cases where the generated comment is completely different. Examples appear in~\cref{tab:example-pos}.

\begin{table*}\centering 
\begin{threeparttable}
  \fontsize{8.6}{11.6}\selectfont
  \begin{tabular}{@{}>{\raggedright}p{0.3\linewidth} | p{0.3\linewidth} |p{0.33\linewidth}@{}}
  \hline
     \textbf{Input Sentence}& \textbf{Gold Output} & \textbf{System Output} \\\hline

    Where there is food \textcolor{red}{\textbf{on}} the smell of cigarette smoke , food soaks Damage to people who go around If you smoke , you know everyone is very often not a good point . & Use the <preposition> `with' instead of <<on>> to express ``food with the smell of cigarettes''. & The <preposition> <<on>> is not used with 'smell' as it is meant to indicate a connection. Use the <preposition> 'in' instead.\\\hline

    The direct means when you smoking , some smoke and chemical access \textcolor{red}{\textbf{to}} your body after that all bad diseases will show the results . & Since the <verb> <<access>> is a <transitive verb>, a <preposition> is not necessary. & <<To>> is not the <preposition> used with the <noun> <<access>> to indicate a point of arrival. Look up the <noun> <<access>> in the dictionary and find the appropriate <preposition>.\\\hline

    And , it is the best \textcolor{red}{\textbf{of}} us to acquire various skills , manners without lazing about study . & <For + noun> should be used as the subject of a <to-infinitive> describing the <formal subject> <<it>>. & <<Of>> is not the <preposition> used with the <adjective> <<best>> to indicate the recipient of an action. Think of a <preposition> that indicates destination.\\\hline

    We are as engineer usually work in an industry or \textcolor{red}{\textbf{on}} a factory , we use our attitude , knowledge to get a job which is comfortable with us . & When the <verb> <<work>> is followed by the <preposition> <<on>>, it means ``to spend time repairing or improving something.'' Use the <preposition> ``in'' or ''at'' to indicate where you work. & It is not incorrect to use the <preposition> <<on>> in this case, but <<on>> describes the state of being physically inside the factory building. Use in' instead to refer to business as a situation or an organization''. \\\hline

  \end{tabular}
  \end{threeparttable}

  \caption{Examples of incorrect system-generated comments. Error span tokens are bold and in red.}
  \label{tab:example}
\end{table*}

\begin{table*}\centering 
\begin{threeparttable}
  \fontsize{8.6}{11.6}\selectfont
  \begin{tabular}{@{}>{\raggedright}p{0.3\linewidth} | p{0.3\linewidth} |p{0.33\linewidth}@{}}
  \hline
     \textbf{Input Sentence}& \textbf{Gold Output} & \textbf{System Output} \\\hline
     
     Smoking sometimes makes people uncomfortable , so smoking should be banned \textcolor{red}{\textbf{at}} the public place like the restaurants .&The <preposition> <<at>> may be used with <<place>> to indicate a place where something takes place, but it is more common to use 'in' instead.&The <preposition> <<at>> may be used with <<place>> to indicate a place where something takes place, but it is more common to use 'in' instead.\\\hline

     In the past , most \textcolor{red}{\textbf{of}} college students probably had no part-time job &A group of something specific follows <<most of>>. Use <<most>> as an <adjective> rather than a <noun> when simply referring to students in general.&<<Most of>> is followed by a <pronoun> or a <noun> with a <determiner> such as 'the', 'these', or'my'. Otherwise, <<most>> comes just before the <noun> to be qualified as an <adjective>.\\\hline

     We meet new people and learn about service for customer , how difficult or hard to get money before we \textcolor{red}{\textbf{graduate school}} . & <<Graduate>> needs to be followed by a <preposition> when used as an <intransitive verb>. & Since the <verb> <<graduate>> is an <intransitive verb>, the <object> requires a <preposition>. Look up the use of the <verb> <<graduate>> in the dictionary. \\\hline

     Parents worry \textcolor{red}{\textbf{about}} they waste time and hurt themselves . & The <preposition> <<about>> follows the <verb> <<worry>> when a <noun> follows. When there is a <clause> in the structure of <S + V>, use a <that-clause>. & The <verb> <<worry>> is a <transitive verb> that takes a <that-clause> as an <object> which does not require a <preposition>. \\\hline

  \end{tabular}
  \end{threeparttable}
  \caption{Examples of correct system-generated comments. Error span tokens are bold and in red.}
  \label{tab:example-pos}
\end{table*}

%% file: 4-discussion.tex
In this paper, we presented a simple baseline for sentence-level feedback generation for English language learners. We investigated the effect of using pseudo data for the task and provided an analysis of system outputs and metrics used.

\paragraph{Does data augmentation help?}
Incorporating pseudo data gave us very slight BLEU score improvements (0.14) over the T5 model with no pseudo data. But in human evaluations, we observe a 3.3\% improvement in F1. Our experiments show that pseudo data could potentially improve results if carefully created and incorporated. Studies on similar tasks such as grammatical error correction have shown that there are many important factors when creating pseudo data such as choice of learner corpora, the method of generating the pseudo data, and error tendency of learners~\cite{kiyono-etal-2019-empirical,white-rozovskaya-2020-comparative,takahashi-etal-2020-grammatical}. Future studies should further investigate similar important factors for this task and their effectiveness.

\paragraph{Is this a general evaluation of grammatical error feedback capability?}
Even though our model outperforms the baseline for this task by a large margin, we think it is not a good indicator of current models' performance for the feedback comment generation task in general, due to the following limitations: 1)~The topics covered in the train/dev/test sets are very limited and only related to smoking and part-time jobs. Considering the topics and the fact that learners may have a limited vocabulary in the target language, the model is likely performing well because it has seen most of the errors during training, not because it has developed the capacity to recognize and comment on grammatical errors in general. 2)~The task is focused on preposition-related errors, which makes the errors and comments even more limited than in realistic settings. 3)~Most of the comments follow a specific template, which made it easier for the model to learn the patterns. 4)~Many reference comments consist of boilerplate---very general suggestions such as \textit{Look up the use of the <verb> <<prohibit>> in a dictionary and rewrite the sentence using the appropriate structure.}
The model also generates many such boilerplate sentences. Of course, a general correct comment is better than a detailed incorrect comment, but we think a model that is able to give more specific suggestions would be of more use to learners.
 
 With these limitations in mind, we still believe this could be a first step toward better and more robust feedback comment generation systems, and we view the organization of the shared task and the release of the data as important milestones for making progress in this research area.